\documentclass{article}

\usepackage{PRIMEarxiv}

\usepackage[utf8]{inputenc} % allow utf-8 input
\usepackage[T1]{fontenc}    % use 8-bit T1 fonts
\usepackage{hyperref}       % hyperlinks
\usepackage{url}            % simple URL typesetting
\usepackage{booktabs}       % professional-quality tables
\usepackage{amsfonts}       % blackboard math symbols
\usepackage{nicefrac}       % compact symbols for 1/2, etc.
\usepackage{microtype}      % microtypography
\usepackage{lipsum}
\usepackage{fancyhdr}       % header
\usepackage{graphicx}       % graphics
\usepackage{amsmath}
\usepackage[table]{xcolor}
\graphicspath{{media/}}     % organize your images and other figures under media/ folder

\title{CAME-AB: Cross-Modality Attention with Mixture-of-Experts for Antibody Binding Site Prediction}

\author{
  Hongzong Li$^*$, \\
  Generative AI Research and Development Center \\
  The Hong Kong University of Science and Technology \\
  Hong Kong\\
  \texttt{lihongzong@ust.hk} \\  
   \And
   Jiahao Ma$^*$, \\
   MILES\\
   The University of Hong Kong\\
   Hetao SZ-HK Cooperation Zone\\
  \texttt{jiahao.ma@connect.hku.hk} \\
   \And
   Zhanpeng Shi\thanks{\textbf{Equal contribution.}}, \\
  College of Veterinary Medicine \\
  Jilin University \\
  Jilin, China\\
  \texttt{shizp9921@mails.jlu.edu.cn} \\
  \And
  Rui Xiao, \\
  School of Chemistry and Chemical Engineering \\
  South China University of Technology \\
  Guangzhou, China\\
  \texttt{202230283066@mail.scut.edu.cn} \\
  \And
  Fanming Jin,\\
  School of Biomedical Sciences \\
  The University of Hong Kong \\
  Hong Kong\\
  \texttt{jinfm@connect.hku.hk} \\
   \And
  Ye-Fan Hu\thanks{%\textit{\dagger}: 
  \textbf{Corresponding authors.}}\,\,,\\
  Computational Immunology Centre \\
  BayVax Biotech Limited \\
  Hong Kong\\
  \texttt{yefan.hu@bayvaxbio.com} \\  
  \And
  Hangjun Che,\\
  College of Electronic and Information Engineering\\
  Southwest University \\
  Chongqing, China\\
  \texttt{hjche123@swu.edu.cn} \\
   \And
  Jian-Dong Huang$^\dagger$\\
  School of Biomedical Sciences \\
  The University of Hong Kong \\
  Hong Kong\\
  \texttt{jdhuang@hku.hk} \\
}

\begin{document}
\maketitle

\begin{abstract}
Antibody binding site prediction plays a pivotal role in computational immunology and therapeutic antibody design. 
%Existing methods often rely on limited sequence-based encodings or experimentally derived structural features, failing to capture the full biological complexity of antibody-antigen interactions.
Existing sequence or structure methods rely on single-view features and fail to identify antibody-specific binding sites on the antigens.
In this paper, we propose \textbf{CAME-AB}, a novel Cross-modality Attention framework with a Mixture-of-Experts (MoE) backbone for robust antibody binding site prediction. CAME-AB integrates five biologically grounded modalities, including raw amino acid encodings, BLOSUM substitution profiles, pretrained language model embeddings, structure-aware features, and GCN-refined biochemical graphs, into a unified multimodal representation. To enhance adaptive cross-modal reasoning, we propose an \emph{adaptive modality fusion} module that learns to dynamically weight each modality based on its global relevance and input-specific contribution. 
A Transformer encoder combined with an MoE module further promotes feature specialization and capacity expansion. We additionally incorporate a supervised contrastive learning objective to explicitly shape the latent space geometry, encouraging intra-class compactness and inter-class separability. To improve optimization stability and generalization, we apply stochastic weight averaging during training.
Extensive experiments on benchmark antibody-antigen datasets demonstrate that CAME-AB consistently outperforms strong baselines on multiple metrics, including Precision, Recall, F1-score, AUC-ROC, and MCC. Ablation studies further validate the effectiveness of each architectural component and the benefit of multimodal feature integration. The model implementation details and the codes are available on \url{https://anonymous.4open.science/r/CAME-AB-C525}
\end{abstract}



\section{Introduction}

Predicting antibody binding site—the regions on antigens recognized by specific antibodies—is a critical task in immunological research, vaccine development, and antibody-based therapeutic design \cite{aguilar2022fragment}. Accurate identification of these binding sites enables a deeper understanding of immune recognition mechanisms, significantly facilitating rational vaccine and therapeutic antibody engineering \cite{parvizpour2020epitope,nagarathinam2024epitope}. Traditional approaches predominantly rely on experimentally determined three-dimensional structures or computational sequence-based methods \cite{liu2020potent}. However, these methods often face limitations due to structural data scarcity and the inability of sequence-based predictors to capture complex structural relationships essential for accurate predictions \cite{fernandez2023challenges}.

Antibody binding site prediction methods based solely on sequence alignment often fail to account for the structural interactions essential to antibody-antigen specificity. For example, sequence-based tools such as BepiPred \cite{clifford2022bepipred}, BLAST \cite{liu2022antibody} and ClustalW \cite{ferrari2021study} are inherently limited in their ability to capture the spatial arrangements and residue interactions critical for accurate prediction. In contrast, structure-based approaches, such as docking simulations and homology modeling, depend heavily on experimentally resolved structures obtained through techniques such as X-ray crystallography or cryo-electron microscopy \cite{wang2017cryo}. However, these methods are both time-intensive and costly, significantly limiting their widespread applicability.

Recent research in computational biology has shifted attention towards integrating multiple data modalities to address these shortcomings \cite{ouyang2024mmsite}. Deep learning methods, particularly transformer architectures and graph neural networks (GNNs), have demonstrated exceptional capability in capturing complex, long-range dependencies in both sequential and spatial data \cite{fu2024learning}. Transformers have shown great potential in language modeling tasks by effectively modeling long-range dependencies through self-attention mechanisms \cite{meng2024gact}. 
Likewise, 
GNNs have demonstrated strong capabilities in capturing spatial dependencies not only within protein tertiary structures but also in modeling the topological relationships in protein–protein interaction \cite{meng2024gact,zhao2023semignn}, thereby showing the importance of incorporating both structural and interaction-based information in predictive tasks \cite{li2024pf2pi, zhao2023semignn}.
The introduction of AlphaFold2 has marked a paradigm shift in protein structure prediction, enabling high-accuracy structural information to be inferred solely from amino acid sequences \cite{jumper2021highly}. The availability of reliable predicted structures provides an unprecedented opportunity to integrate structural context without requiring experimental resolution, significantly expanding the practical applicability of computational predictions.
Additionally, protein language models, such as ProteinBERT \cite{brandes2022proteinbert} and evolutionary scale modeling (ESM) \cite{israeli2024single}, have demonstrated impressive capability in encoding evolutionary and functional information into dense continuous embeddings. These pretrained models have captured deep evolutionary relationships, contributing significantly to the performance of downstream prediction tasks. Integrating these pretrained embeddings into epitope prediction tasks offers a promising strategy to enhance prediction accuracy and generalizability.

\begin{figure}[htbp]
	\centering
	\includegraphics[width=0.55\linewidth]{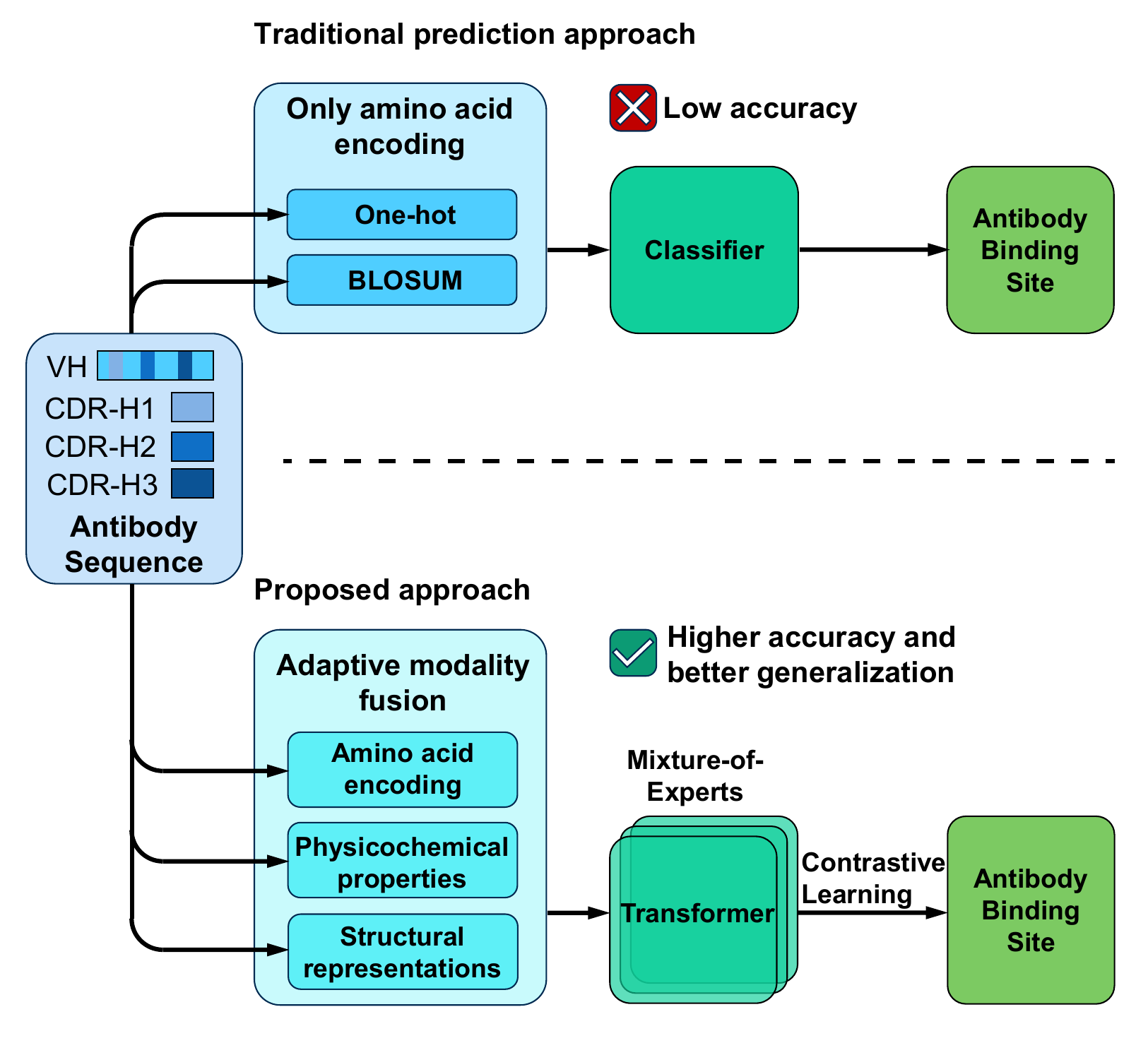}
	\caption{Comparison between the traditional prediction approach and our proposed cross-modality attention learning framework.}
	\label{fig:intro}
\end{figure}

Motivated by recent advances in protein modeling and multimodal learning, we propose \textbf{CAME-AB}, a novel Cross-Modality Attention framework equipped with a Mixture-of-Experts (MoE) backbone for antibody binding site prediction. CAME-AB systematically integrates complementary biological information from multiple representation spaces (as shown in Figure~\ref{fig:intro}). Given the critical role of the CDR region in antigen-antibody recognition~\cite{jones1986replacing}, CAME-AB focuses on the antibody heavy-chain variable region (VH), including CDR-H1, CDR-H2, and CDR-H3 loops, and extracts semantically rich features using diverse encoding strategies.
Specifically, CAME-AB incorporates five biologically grounded modalities: (i) one-hot encoding and BLOSUM matrices to capture residue identity and evolutionary substitution patterns; (ii) pretrained contextual embeddings from a large protein language model, i.e., ESMC; (iii) structural features derived from ESM’s structure-aware output layers; and (iv) residue-level biochemical similarity graphs constructed using PyBioMed descriptors, from which we obtain structural-aware node embeddings via a graph convolutional network. These heterogeneous representations are projected into a unified latent space and fused via a learnable adaptive modality fusion module that jointly models modality informativeness, sample-specific variation, and class-aware semantic priors.
To improve discriminative capacity and model generalization, our proposed architecture incorporates three additional components: (1) a Mixture-of-Experts (MoE) module to encourage feature specialization across latent subspaces; (2) a supervised contrastive learning objective to enforce intra-class compactness and inter-class separability in the embedding space; and (3) Stochastic Weight Averaging (SWA) for optimization smoothing and enhanced generalization.
We evaluate our framework on multiple public antibody-antigen binding datasets. Extensive experiments show that our method consistently outperforms state-of-the-art baselines on multiple metrics. Ablation studies further verify the effectiveness of each multi-view feature and architectural component.

\textbf{Our key contributions are summarized as follows:}
\begin{itemize}
    \item We present a unified multimodal deep learning framework that integrates sequence, structural, and biochemical modalities for antibody binding site prediction.
    \item We propose a novel combination of adaptive modality fusion, contrastive learning, and MoE, jointly optimized under a robust training strategy incorporating SWA.
    \item We achieve state-of-the-art performance on benchmark datasets and provide comprehensive ablation studies to validate the contribution of each design component.
\end{itemize}

%The remainder of this paper is structured as follows. Section 2 provides an overview of related work, including \lhz{epitope prediction}, multiview representation, transformer, mixture-of-experts, contrastive learning loss, and stochastic weight averaging. Section 3 presents our multimodal transformer-based model in detail, including feature engineering strategies. Section 3 describes the experimental setup and validation procedures. Section 4 presents comprehensive results, including quantitative comparisons, ablation studies, and qualitative analyses. Finally, Section 5 concludes with a summary of contributions and outlines directions for future research.

\section{Related Work}

\subsection{Antibody Binding Site Prediction}

Antibody binding site prediction aims to identify antigen regions (epitopes) capable of interacting with antibodies. While critical for vaccine design and therapeutic development, conventional approaches exhibit a fundamental limitation: they predict \textit{where} binding may occur on an antigen surface, but cannot determine \textit{which specific antibodies} would recognize these epitopes. This distinction is crucial for developing targeted immunological interventions. %\hyf{Since all the antigens in our dataset are viral proteins, antibodies capable of binding to them must exist—the main challenge is determining which antibody can bind. \textbf{This sentence can be placed somewhere else}} 

Existing methods fall into two categories with inherent constraints:
\begin{itemize}
\item \textbf{Sequence-based models} (e.g., BepiPred~\cite{clifford2022bepipred}, ABCPred~\cite{malik2022abcpred}) employ machine learning on sequence features to predict linear epitopes. Although computationally efficient, they ignore spatial context and fail to capture conformational epitopes.

\item \textbf{Structure-based methods} (e.g., Molecular Docking~\cite{gaudreault2023enhanced}) utilize 3D structural information through docking simulations and geometric analysis. Although better at identifying spatial epitopes, these approaches require experimentally resolved structures that are often unavailable.

\end{itemize}

Notably, both paradigms share a critical shortcoming: they generate \textit{generic} epitope predictions without antibody-specific binding information. A predicted epitope region might theoretically bind multiple antibody clones, but existing methods cannot discriminate which specific pairing of paratope-epitope would occur in practice.

%The emergence of AlphaFold2~\cite{yang2023alphafold2} presents new opportunities to bridge this gap. By enabling accurate \textit{de novo} structure prediction from sequences, it allows systematic integration of structural and sequential features at scale. This advancement paves the way for next-generation models that could potentially incorporate antibody-specific characteristics (e.g., CDR loop geometry) to predict \textit{specific} binding interactions rather than generic epitope locations.

\subsection{Feature Representation in Bioinformatics}

Feature representation is a cornerstone of bioinformatics, enabling the extraction and integration of meaningful patterns from biological data. In this section, we introduce key feature representation methods, including ESMC, One-hot encoding, BLOSUM \cite{mount2008using}, and PyBioMed \cite{dong2018pybiomed}, alongside a discussion of multi-view learning approaches and their relevance to bioinformatics tasks.
A detailed description of each feature representation and its bioinformatics relevance is provided in the Supplementary Material (Section~\ref{fea_rep}).

\section{Methodology}

% Overview of our proposed adaptive multimodal transformer framework for antibody binding site prediction. The model integrates five biologically grounded modalities extracted from antibody heavy-chain VH region sequences: one-hot encoding, BLOSUM matrices, pretrained protein language embeddings (e.g., ESMC), structure-informed embeddings from ESMC’s structure head, and GCN-derived biochemical features. Each modality is independently projected into a shared 256-dimensional latent space while preserving its modality-specific semantics. An Adaptive Modality Fusion (AMF) module then computes a fused representation by dynamically weighting each modality based on global informativeness, sample-level variation, and class-aware semantics. The fused representation is processed by a two-layer Transformer encoder to capture intra-sequence residue interactions, followed by a Mixture-of-Experts (MoE) module to enhance specialization. The resulting expert-refined embedding is simultaneously used for classification and contrastive learning, where a projection head encourages intra-class compactness and inter-class separability via supervised contrastive loss. Stochastic Weight Averaging (SWA) is employed to improve training stability and generalization.

\subsection{Problem Formulation and Multimodal Representation}

The objective of this work is to predict the antigen epitope binding class based on antibody heavy-chain sequences, focusing on the VH region and its complementarity-determining regions (i.e., CDR-H1, CDR-H2, and CDR-H3). Existing sequence-based models often suffer from limited representation capacity, failing to capture the full spectrum of biochemical, evolutionary, and structural information required for accurate epitope recognition.

\begin{figure}[htbp]
	\centering
	\includegraphics[width=\linewidth]{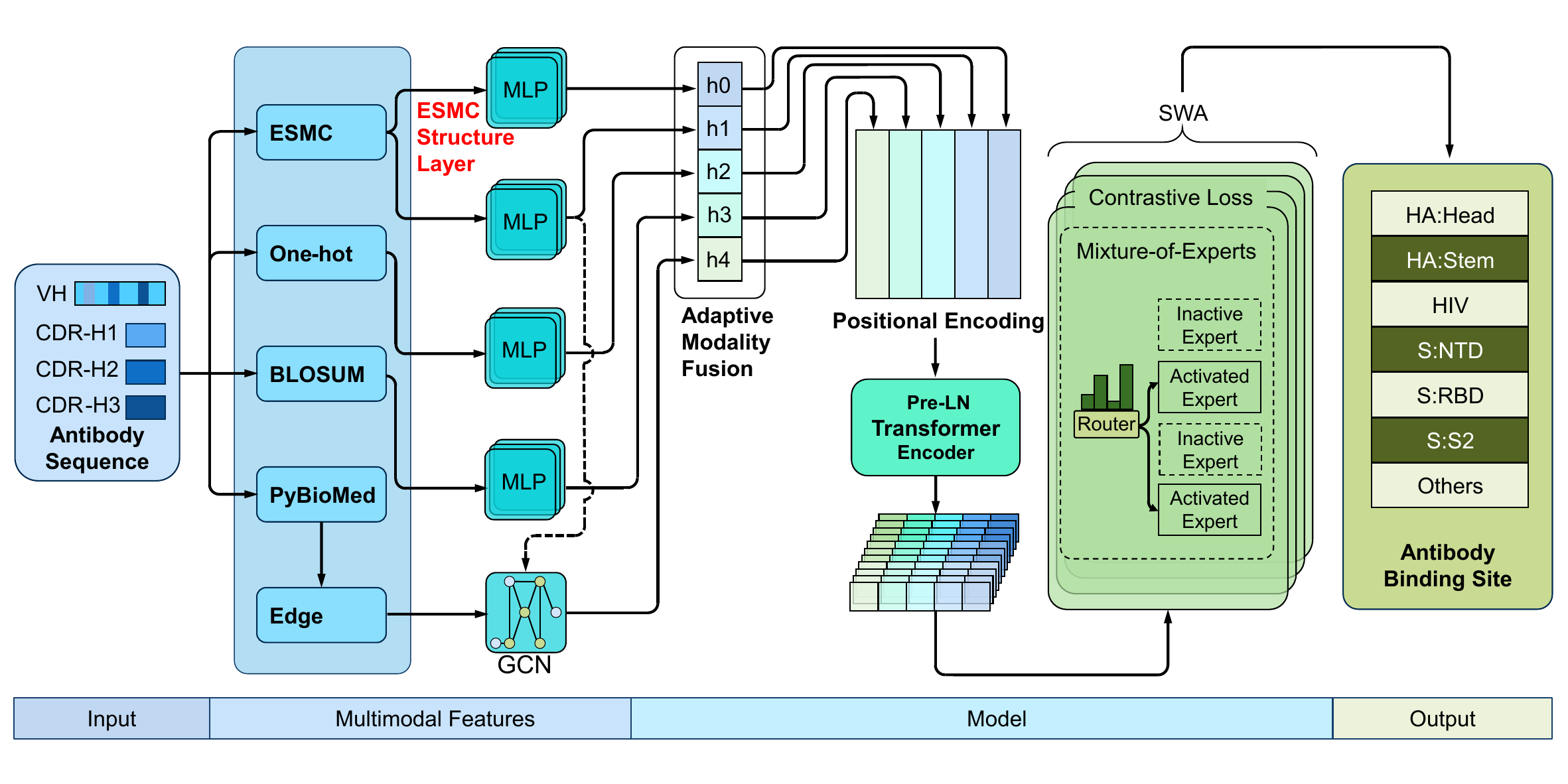}
	\caption{
    Overview of our adaptive multimodal transformer for antibody binding site prediction. The model integrates five modalities from antibody VH sequences: one-hot encoding, BLOSUM, pretrained protein embeddings (ESMC), structure-informed embeddings, and GCN-based biochemical features. Each modality is projected into a shared 256-dimensional space. An Adaptive Modality Fusion (AMF) module dynamically weights and fuses these modalities. The fused representation is processed by a Transformer encoder to capture residue interactions, followed by a Mixture-of-Experts (MoE) for specialization. The final embedding supports both classification and contrastive learning, promoting class discrimination. Stochastic Weight Averaging (SWA) enhances training stability and generalization.
}
	\label{fig:model_overview}
\end{figure}

To address this, we formulate epitope prediction as a multimodal learning task. Our framework integrates five biologically grounded feature modalities as shown in Figure \ref{fig:model_overview}: 
(i) \textbf{Amino acid encoding schemes:} we incorporate one-hot encoding to preserve raw residue identity, BLOSUM substitution matrices to model evolutionary conservation patterns, and contextualized embeddings derived from pretrained protein language models such as ESMC to capture sequence semantics and long-range dependencies; 
(ii) \textbf{Structure-informed representations:} we utilize the structural output layer of ESMC as a dedicated structural modality. This layer provides an approximate estimation of spatial residue relationships even in the absence of experimentally resolved structures; 
(iii) \textbf{Graph-based biochemical features:} we construct a residue-level graph where each node corresponds to an amino acid and is initialized using its ESMC sequence embedding. Edges are established between residue pairs based on pairwise biochemical similarity, computed from PyBioMed-derived physicochemical descriptors such as hydrophobicity, polarity, and surface accessibility. This graph is used to train a Graph Convolutional Network (GCN) \cite{kipf2017semi}, for refining node embeddings by aggregating spatial and chemical context from neighboring residues.

Let \( F_i \) denote the aggregated multimodal representation of the $i$-th antibody sample. Our model aims to learn a predictive function \( f(F_i; \theta) \), where \( \theta \) denotes learnable parameters, such that:

\begin{equation}
	\hat{y}_i = f(F_i; \theta), \quad \hat{y}_i \in \{1, 2, \ldots, C\},
\end{equation}
where \( \hat{y}_i \) is the predicted epitope class and \( C \) is the total number of classes. This multimodal formulation enables holistic modeling of antibody properties, improving generalization and robustness.

\subsection{Architecture Overview}

As shown in Figure \ref{fig:model_overview}, our proposed deep learning architecture addresses the challenges of multimodal integration, representation specialization, and inter-class discrimination. It consists of four key components: 
(1) \textbf{Multimodal feature encoding}; 
(2) \textbf{Adaptive modality fusion}; 
(3) \textbf{Transformer-based backbone with Mixture-of-Experts (MoE)}; 
(4) \textbf{Prediction and Contrastive Embedding}.

\subsubsection{Multimodal Feature Encoding}

To construct unified representations, we first process each feature modality independently. Specifically, we extract the following biologically grounded features for each antibody sequence:

\begin{itemize}
    \item \textbf{One-hot encoding} ($F^{\text{onehot}} \in \mathbb{R}^{L \times 20}$): Encodes discrete residue identities across the sequence.
    \item \textbf{BLOSUM features} ($F^{\text{blosum}} \in \mathbb{R}^{L \times 20}$): Capture residue-level substitution propensities from evolutionary matrices.
    \item \textbf{ESMC embeddings} ($F^{\text{esm}} \in \mathbb{R}^{L \times d_1}$): Contextualized token embeddings derived from pretrained language models such as ESM or ProteinBERT, encoding semantic and evolutionary context.
    \item \textbf{Structure-aware embeddings} ($F^{\text{struct}} \in \mathbb{R}^{L \times d_2}$): Extracted from the structure-specific output layer of ESMC models, reflecting residue spatial characteristics inferred from AlphaFold2-style estimators.
    \item \textbf{GCN-based physicochemical embeddings} ($F^{\text{gcn}} \in \mathbb{R}^{L \times d_3}$): Computed by applying a GCN to a residue-level graph that encodes biochemical similarities.
\end{itemize}

For the GCN branch, we construct a residue-level graph \( G = (V, E) \), where each node \( v_i \in V \) corresponds to a residue and is initialized using its ESMC embedding. Edges \( (v_i, v_j) \in E \) are established based on pairwise biochemical similarity, computed using PyBioMed-derived descriptors such as hydrophobicity, polarity, and charge. residues are connected if their pairwise similarity, computed using selected PyBioMed descriptors, exceeds a threshold.
Given the graph \( G \), we apply a two-layer GCN \cite{kipf2017semi} to refine the node embeddings. Let \( \mathbf{X} \in \mathbb{R}^{L \times d} \) be the input node feature matrix and \( \mathbf{A} \in \mathbb{R}^{L \times L} \) be the adjacency matrix of the graph \( G \), the GCN layer is defined as \cite{kipf2017semi}:
\begin{equation}
\mathbf{H}^{(l+1)} = \sigma\left( \hat{\mathbf{D}}^{-\frac{1}{2}} \hat{\mathbf{A}} \hat{\mathbf{D}}^{-\frac{1}{2}} \mathbf{H}^{(l)} \mathbf{W}^{(l)} \right),
\end{equation}
where \( \hat{\mathbf{A}} = \mathbf{A} + \mathbf{I} \) is the adjacency matrix with self-loops, \( \hat{\mathbf{D}} \) is the diagonal degree matrix of \( \hat{\mathbf{A}} \), \( \mathbf{H}^{(0)} = \mathbf{X} \), \( \mathbf{W}^{(l)} \) is the trainable weight matrix of layer \( l \), and \( \sigma(\cdot) \) is a non-linear activation function (e.g., ReLU or GELU).
After two propagation steps, the final output \( \mathbf{H}^{(2)} \in \mathbb{R}^{L \times d'} \) is treated as the GCN-based structural modality \( F^{\text{gcn}} \) in our framework. This modality captures residue-level spatial and biochemical context, complementing sequence-derived representations.

Each feature matrix \( F^{(m)} \in \mathbb{R}^{L \times d_m} \), where $m = 1, \dots, M$, is projected into a shared latent space of dimension \( d = 256 \) via a modality-specific transformation:

\begin{equation}
    \tilde{F}^{(m)} = \text{Dropout}\left(\text{GELU}\left(\text{LayerNorm}\left(F^{(m)} W^{(m)} + b^{(m)}\right)\right)\right),
\end{equation}
where \( W^{(m)} \in \mathbb{R}^{d_m \times d} \), \( b^{(m)} \in \mathbb{R}^d \) are learnable parameters for each modality.

This modular encoding strategy ensures that the distinctive semantics of each modality are preserved prior to fusion. The resulting set of aligned representations \( \{\tilde{F}^{(m)}\}_{m=1}^M \in \mathbb{R}^{L \times d} \) is forwarded to the adaptive modality fusion module for cross-view interaction learning.

\subsubsection{Adaptive Modality Fusion}

To effectively integrate diverse biological modalities while preserving their respective contributions, we propose an Adaptive Modality Fusion (AMF) module. Unlike naïve concatenation or fixed-weight averaging, our AMF module learns to dynamically assign weights to each modality based on global importance, sample-specific variation, and class-aware semantics.

Inspired by hierarchical gating and label-aware conditioning to mitigate spurious modality correlations~\cite{li2025devil}, our approach introduces three types of adaptive weights.
Let $\tilde{F}^{(m)} \in \mathbb{R}^{L \times d}$ denote the projected embedding of the $m$-th modality. The fused representation $\mathbf{F}_{\text{fused}} \in \mathbb{R}^{L \times d}$ is computed as a weighted sum over all $M$ modalities:

\begin{equation}
\mathbf{F}_{\text{fused}} = \sum_{m=1}^{M} \alpha_m \cdot \beta^{(i)}_m \cdot \gamma^{(y_i)}_m \cdot \tilde{F}^{(m)},
\end{equation}
where $\alpha_m \in [0,1]$ is a learnable global importance score for modality $m$, $\gamma^{(y_i)}_m$ is a class-aware weight dependent on the ground truth epitope class $y_i$, implemented as a learnable embedding lookup: $\gamma^{(y_i)}_m = \text{Embed}(y_i)[m]$, and $\beta^{(i)}_m$ is a sample-specific weight computed via a gating network as follows:
\begin{equation}
\beta^{(i)}_m = \text{softmax}_m\left(W_{\beta} \cdot \text{Pool}(\tilde{F}^{(m)}_i) + b_{\beta}\right),
\end{equation}
where $\text{Pool}(\cdot)$ applies mean pooling across residues, and $W_{\beta}, b_{\beta}$ are learnable.

This triple-weight mechanism enables the model to:
(1) emphasize universally informative modalities;
(2) adapt to individual antibody input profiles;
(3) condition integration on task-specific class semantics.
All weights are jointly optimized during training through backpropagation, encouraging end-to-end alignment across modalities and output space.

\subsubsection{Transformer-based Representation Encoding with Mixture-of-Experts}

After adaptive fusion, the integrated representation $\mathbf{F}_{\text{fused}} \in \mathbb{R}^{L \times d}$ is passed through a two-layer Transformer encoder to capture intra-sequence dependencies and inter-residue interactions across CDRs. We adopt a Pre-LayerNorm (Pre-LN) architecture \cite{xiong2020layer} for improved training stability, defined as:

\begin{align}
\mathbf{H}_0 &= \mathbf{F}_{\text{fused}}, \\
\mathbf{A}_l &= \mathbf{H}_{l-1} + \text{MHSA}(\text{LayerNorm}(\mathbf{H}_{l-1})), \\
\mathbf{H}_l &= \mathbf{A}_l + \text{FFN}(\text{LayerNorm}(\mathbf{A}_l)), \quad l = 1, ..., n,
\end{align}
where $\text{MHSA}(\cdot)$ denotes multi-head self-attention and $\text{FFN}(\cdot)$ is a feedforward sublayer with GELU activation and dropout. The output $\mathbf{H}_2$ is mean-pooled across sequence length to obtain a condensed representation $\mathbf{z}_i \in \mathbb{R}^d$ for each antibody.

To further enhance feature specialization and model capacity, we introduce a Mixture-of-Experts (MoE) module \cite{shazeer2017outrageously}. It consists of $K$ expert networks $\{E_k\}_{k=1}^{K}$, each implemented as a two-layer MLP. A gating network assigns a soft distribution over experts \cite{shazeer2017outrageously}:
\begin{align}
\mathbf{g} &= \text{softmax}(W_g \cdot \mathbf{z}_i + b_g), \\
\mathbf{h}_{\text{moe}} &= \sum_{k=1}^{K} g_k \cdot E_k(\mathbf{z}_i),
\end{align}
where \( W_g \in \mathbb{R}^{d \times K} \) is the gating weight matrix and \( b_g \in \mathbb{R}^K \) is a learnable bias vector. The gating network transforms the fused representation \( \mathbf{z}_i \) into a soft distribution \( \mathbf{g} \in \mathbb{R}^K \) over \( K \) expert modules. The bias term \( b_g \) adjusts the prior logarithmics of each expert before applying softmax, allowing the model to learn a global preference or offset for each expert regardless of the input sample. This is critical when certain experts are more generally informative or require activation even under low attention from the gating vector. The final expert-refined embedding \( \mathbf{h}_{\text{moe}} \) is computed as a weighted sum over all expert outputs \( E_k(\mathbf{z}_i) \), enabling dynamic specialization across the expert ensemble.

To prevent expert collapse and encourage diverse specialization, we introduce a diversity regularization loss \cite{kim2019diversify}:

\begin{equation}
\mathcal{L}_{\text{diversity}} = \frac{1}{K(K-1)} \sum_{i=1}^{K} \sum_{j\neq i} \text{cos\_sim}(E_i, E_j),
\end{equation}
where $\text{cos\_sim}(\cdot,\cdot)$ computes cosine similarity between expert outputs. This encourages experts to focus on complementary feature subspaces and improves model robustness.

\subsubsection{Prediction and Contrastive Embedding}

The MoE-refined embedding $\mathbf{h}_{\text{moe}}$ is used for classification, uncertainty estimation, and contrastive representation learning. A two-layer MLP classifier outputs the logits:

\begin{equation}
\hat{y} = \text{MLP}_{\text{cls}}(\mathbf{h}_{\text{moe}}) = W_4 (\text{GELU}( \text{LN}(W_3 \mathbf{h}_{\text{moe}} + b_3))) + b_4.
\end{equation}

Although the classifier provides direct supervision via focal loss, we introduce a contrastive learning objective to explicitly regularize the geometry of the latent space. This objective promotes \textit{intra-class compactness} and \textit{inter-class separability}, enhancing the robustness and generalizability of learned representations.

To this end, we add a projection head that maps the expert-refined representation \( \mathbf{h}_{\text{moe}}^{(i)} \in \mathbb{R}^{d} \) for sample \( i \) into a contrastive embedding space:

\begin{equation}
\mathbf{z}_i = W_2 \left( \text{GELU}\left( W_1 \cdot \mathbf{h}_{\text{moe}}^{(i)} + b_1 \right) \right) + b_2,
\end{equation}
where \( W_1, W_2 \) are trainable projection matrices and \( \mathbf{z}_i \in \mathbb{R}^{d'} \) is the projected embedding. All embeddings \( \{ \mathbf{z}_i \} \) are normalized to unit length before similarity calculation.

Let \( P(i) \subseteq \{1, \ldots, N\} \) denote the set of indices of positive samples in the batch that share the same ground-truth label as sample \( i \), and let \( A(i) = \{1, \ldots, N\} \setminus \{i\} \) be the set of all other samples excluding \( i \) itself. The supervised contrastive loss is then defined as \cite{khosla2020supervised}:
\begin{equation}
\mathcal{L}_{\text{contrast}} = \sum_{i=1}^{N} \frac{-1}{|P(i)|} \sum_{p \in P(i)} \log \frac{\exp\left(\text{com\_sim}(\mathbf{z}_i, \mathbf{z}_p)/\tau\right)}{\sum_{a \in A(i)} \exp\left(\text{com\_sim}(\mathbf{z}_i, \mathbf{z}_a)/\tau\right)},
\end{equation}
where $\mathbf{z}_p$ is the contrastive embedding of a positive sample $p \in P(i)$, $\text{com\_sim}(\cdot, \cdot)$ denotes cosine similarity, and $\tau$ is a temperature hyperparameter.

This formulation encourages each sample's representation \( \mathbf{z}_i \) to be close to other embeddings of the same class \( \mathbf{z}_p \), while pushing it away from those of other classes. In practice, we further improve discriminative capacity by applying hard negative mining, class-aware sampling, and feature-level augmentation strategies.

%\lhz{We employ hard negative mining, class-weighted sampling, and feature-level augmentations (e.g., noise injection, modality mixup, CutMix) to enrich representation learning.}

% In addition, each modality-specific embedding contributes to auxiliary classification losses for better disentanglement. We also employ an uncertainty head to estimate the confidence of predictions:

% \begin{equation}
% u = \sigma(W_u^{(2)} \cdot \text{GELU}(W_u^{(1)} \cdot \mathbf{h}_{\text{moe}})),
% \end{equation}

% where $\sigma(\cdot)$ is the sigmoid function. The uncertainty prediction can be used to downweight noisy examples or identify ambiguous cases for downstream filtering.

To stabilize training, we apply Stochastic Weight Averaging (SWA)~\cite{zhang2020swa} during the final training phase:

\begin{equation}
\theta_{\text{SWA}} \leftarrow \frac{1}{t - S + 1} \sum_{i=S}^{t} \theta_i.
\end{equation}

The focal loss function was originally proposed to address class imbalance by down-weighting easy examples and focusing learning on hard, misclassified samples, as defined as~\cite{lin2017focal}:
\begin{equation}
\mathcal{L}_{\text{focal}} = -\alpha_y (1 - p_y)^\gamma \log(p_y),
\end{equation}
where \( p_y \) is the predicted probability for the ground-truth class \( y \), \( \alpha_y \in [0,1] \) is a class-balancing weight, \( \gamma \ge 0 \) is a focusing parameter. The modulating term \( (1 - p_y)^\gamma \) dynamically scales the standard cross-entropy loss. When \( p_y \) is high (i.e., the prediction is confident and correct), the term is small, reducing the loss contribution from well-classified samples. Conversely, when \( p_y \) is low, the loss is amplified, emphasizing challenging or underrepresented instances.

Our final training objective combines multiple loss components:
\begin{equation}
\mathcal{L}_{\text{total}} = \mathcal{L}_{\text{focal}} + \lambda_{\text{aux}} \cdot \mathcal{L}_{\text{modal}} + \lambda_{\text{contrast}} \cdot \mathcal{L}_{\text{contrast}} + \lambda_{\text{div}} \cdot \mathcal{L}_{\text{diversity}},
\end{equation}
where the focal loss \( \mathcal{L}_{\text{focal}} \) enhances robustness to imbalanced data, the scalar weights \( \lambda_{\text{aux}}, \lambda_{\text{contrast}}, \lambda_{\text{div}} \) are hyperparameters that balance the auxiliary modality losses, supervised contrastive loss, and expert diversity regularization, respectively.

\section{Experiments}

\subsection{Experimental Setup}
\subsubsection{Dataset and Preprocessing}

To ensure fair evaluation and reproducibility, we construct dataset following a standardized pipeline in ABS~\cite{wang2024explainable}. Specifically, we collect antibody sequences annotated with antigen-binding information, focusing on the VH, CDR1, CDR2, and CDR3 regions. Sequences containing missing or incomplete entries are discarded to ensure data integrity.
To mitigate redundancy and reduce potential data leakage, we perform sequence clustering using CD-HIT \cite{fu2012cd} at a 90\% identity threshold. Clusters are further refined using an 80\% similarity threshold   \cite{wang2002clustering} to construct non-overlapping training, validation, and test partitions. This stratification guarantees that highly similar sequences do not appear across different splits, thereby promoting a robust generalization evaluation.
To address class imbalance, we apply a combination of noise reduction and controlled up/down-sampling strategies within each split. The resulting dataset is divided into 80\% for training, 10\% for validation, and 10\% for testing. Additionally, we remove all duplicate entries post-clustering to ensure that each sequence instance contributes uniquely to model training and evaluation.

\subsubsection{Implementation Details}

Our model is implemented in PyTorch and optimized using the Adam optimizer. We adopt a cyclical learning rate schedule with an initial learning rate of \(10^{-4}\), decaying exponentially by 0.95 every 10 epochs. Training is conducted for 50 epochs with early stopping based on validation loss. The batch size is set to 64.

Hyperparameter tuning is performed via grid search over embedding dimensions \{64, 128, 256\}, transformer depth \{2, 4, 6\}, and attention heads \{4, 8, 12\}. Dropout is applied at 0.1 rate for all layers, and weight decay is set to \(10^{-5}\).

For evaluation, we adopt five standard metrics—Precision, Recall, F1-score, AUC-ROC, and Matthews Correlation Coefficient (MCC)—to comprehensively assess model performance. The definitions of these metrics are provided in the Supplementary Material (Section 2).

\subsection{Ablation Studies}

To quantify the contribution of different input modalities and architectural components, we perform systematic ablations on our full model.

\subsubsection{Impact of Input Modalities}

Table~\ref{tab:ablation_input} reports the performance when each modality is removed. Excluding the ESMC pretrained embeddings yields the largest performance drop, reducing F1-score by 1.94\% and MCC by 0.0307, highlighting the critical role of contextualized protein language features. Removing BLOSUM causes a sharp decline in recall (-17.5\%), suggesting its importance in capturing evolutionary substitution patterns.

GCN-based features provide modest yet consistent gains, showing their complementary role in encoding physicochemical spatial dependencies. The removal of one-hot encoding leads to uniform degradation, validating its utility despite being a low-level encoding.

\begin{table}[htbp]
    \centering
    \caption{Ablation results: performance impact of removing different input feature modalities.}
    \label{tab:ablation_input}
    \resizebox{\textwidth}{!}{
	\begin{tabular}{lccccc}
		\toprule
		Feature Set Removed & Precision & Recall & F1-score & AUC-ROC & MCC \\
		\midrule
		\rowcolor{gray!40} Full Model & \textbf{0.8227±0.0031}& \textbf{0.8250±0.0019}& \textbf{0.8185±0.0021}& 0.9351±0.0025& \textbf{0.7134±0.0030}
		\\
		w/o ESMC & 0.8013±0.0061& 0.8072±0.0025& 0.7991±0.0025& 0.9223±0.0041& 0.6827±0.0060
		\\
		w/o ESMC Structure & 0.8097±0.0043& 0.8093±0.0043& 0.8014±0.0029& \textbf{0.9367± 0.0033}& 0.6900±0.0052
		\\
		w/o One-hot & 0.8148±0.0047& 0.8182±0.0018& 0.8138±0.0039& 0.9361±0.0036& 0.7038±0.0053
		\\
		w/o BLOSUM & 0.8156±0.0123& 0.6494±0.0092& 0.7021± 0.0065& 0.912±0.0143& 0.5226±0.0064
		\\
		w/o GCN &  0.8223±0.0040& 0.8199±0.0043& 0.8147±0.0036& 0.9363±0.0044& 0.7056± 0.0071
		\\
		\bottomrule
	\end{tabular}
    }
\end{table}

\subsubsection{Impact of Model Components}

Table~\ref{tab:ablation_arch} shows the impact of removing architectural modules. Disabling adaptive modality fusion (AMF) results in noticeable performance degradation, particularly in F1-score (-0.92\%) and MCC (-0.0144), underscoring the benefit of class-aware dynamic fusion.

Contrastive learning slightly boosts all metrics by enhancing class-level separability, while removing the MoE block significantly reduces recall, highlighting its contribution to expert-level specialization. Removing SWA leads to the best AUC but lower MCC, suggesting that SWA benefits generalization rather than decision boundary sharpness.

\begin{table*}[htbp]
    \centering
    \caption{Ablation results: effect of removing key architectural modules.}
    \label{tab:ablation_arch}
    \resizebox{\textwidth}{!}{
	\begin{tabular}{lccccc}
		\toprule
		Architectural Module Removed & Precision & Recall & F1-score & AUC-ROC & MCC \\
		\midrule
		\rowcolor{gray!40} Full Model & \textbf{0.8227±0.0031}& \textbf{0.8250±0.0019}& \textbf{0.8185±0.0021}& 0.9351±0.0025&\textbf{ 0.7134±0.0030}
		\\
		w/o contrastive learning & 0.8145±0.0033& 0.8197±0.0023& 0.8123±0.0033& 0.9394±0.0053& 0.7035±0.0048
		\\
		w/o adaptive modal fusion & 0.8138±0.0023& 0.8196±0.0008& 0.8127±0.0022& 0.9343±0.0026& 0.7033±0.0025
		\\
		w/o MoE& 0.8158±0.0033& 0.8177±0.0014& 0.8118±0.0013& \textbf{0.9440±0.0005}&0.7016±0.0017\\
		w/o SWA & 0.8147±0.0017& 0.8184±0.0023& 0.8113±0.0015&0.9397±0.0049& 0.7020±0.0023
		\\
		\bottomrule
	\end{tabular}
    }
\end{table*}

\subsection{Comparison with Baselines}

\begin{figure}[htbp]
    \centering
    \includegraphics[width=0.5\linewidth]{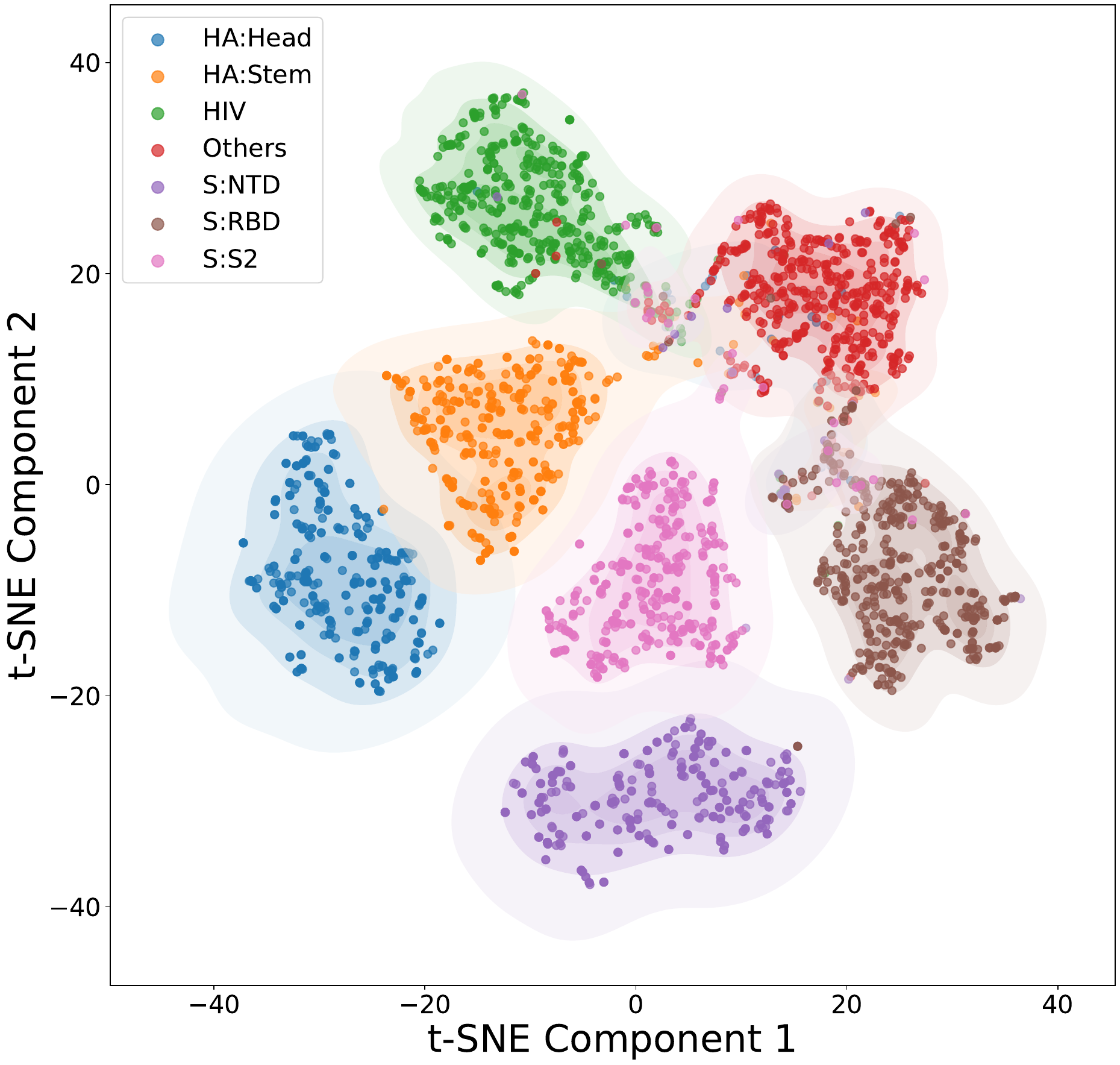}
    \caption{t-SNE visualization of learned feature embeddings. Our model exhibits strong intra-class compactness and inter-class separability, showing the effectiveness of contrastive supervision and multimodal fusion.}
    \label{fig:tsne}
\end{figure}
\begin{figure}[htbp]
    \centering
    \includegraphics[width=0.5\linewidth]{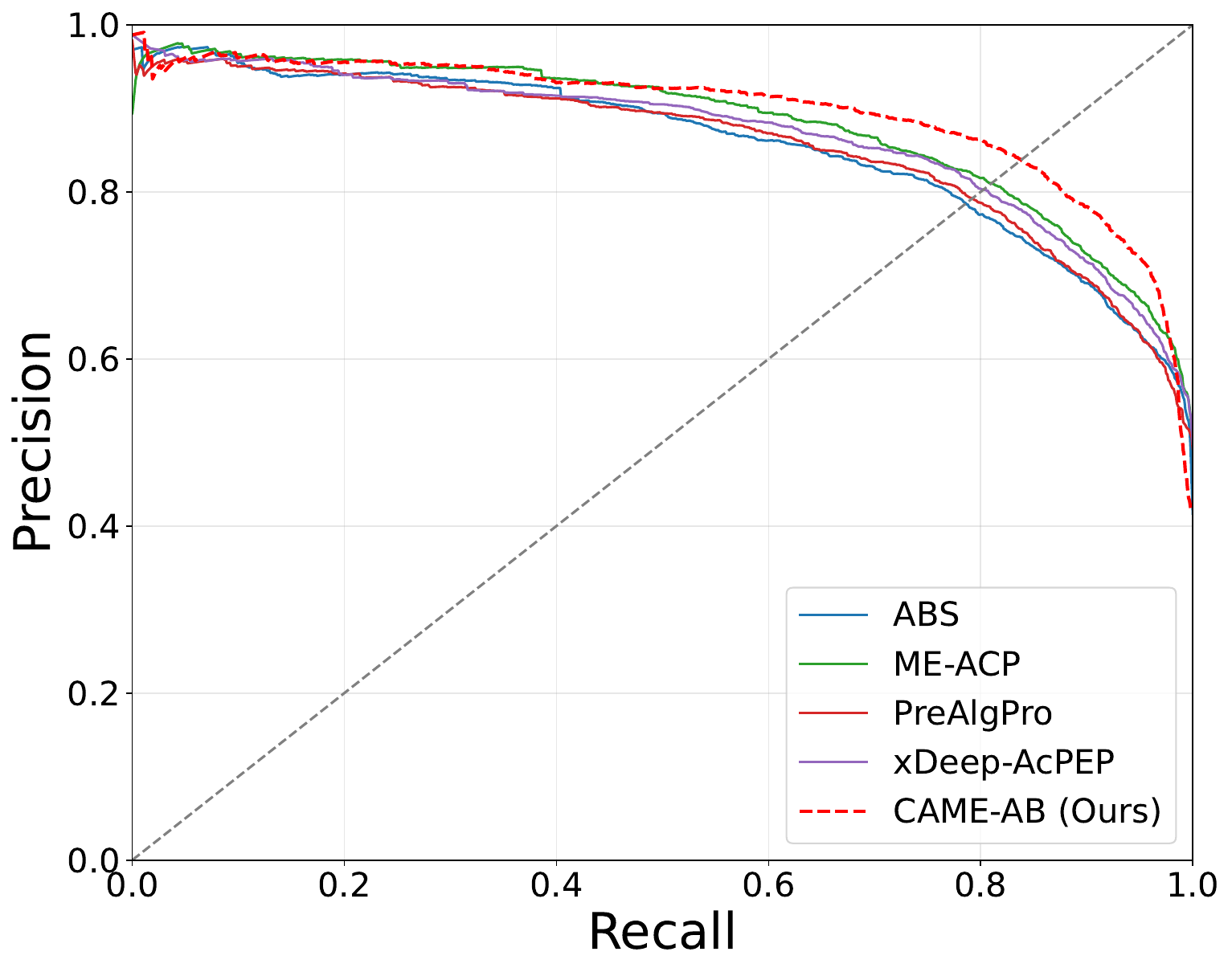}
    \caption{Precision-Recall curve comparison (micro-averaged). The proposed method maintains consistently high precision across all recall levels, indicating superior calibration and prediction reliability.}
    \label{fig:pr_curve}
\end{figure}

To evaluate the effectiveness of our proposed approach, we compare it against several state-of-the-art (SOTA) methods, including ABS~\cite{wang2024explainable}, ME-ACP~\cite{feng2022me}, xDeep-AcPEP~\cite{chen2021xdeep}, and PreAlgPro~\cite{zhang2024prealgpro}.

\begin{table*}[htbp]
    \centering
    \caption{Comparison with state-of-the-art methods.}
    \label{tab:comparison_results}
    \resizebox{\textwidth}{!}{
	\begin{tabular}{lccccc}
		\toprule
		Model & Precision & Recall & F1-score & AUC-ROC & MCC \\
		\midrule
		ABS~\cite{wang2024explainable}& 0.8041±0.0024& 0.6934±0.0025& 0.7336±0.0024& 0.9316±0.0004& 0.5528±0.0037
		\\
		ME-ACP~\cite{feng2022me} & 0.8122±0.0031& 0.8198±0.0026& 0.8140±0.0029& \textbf{0.9587±0.0005}& 0.7035±0.0044
		\\
		xDeep-AcPEP~\cite{chen2021xdeep}& 0.7944±0.0126& 0.8023±0.0118&  0.7955±0.0105& 0.9093±0.0067& 0.6754±0.0180
		\\
		PreAlgPro~\cite{zhang2024prealgpro}& 0.7754±0.0112& 0.7756±0.0071& 0.7749±0.0085& 0.9332±0.0023& 0.6354±0.0125
		\\
		\rowcolor{gray!40}\textbf{CAME-AB (herein)} &\textbf{ 0.8227±0.0031}& \textbf{0.8250±0.0019}& \textbf{0.8185±0.0021}& 0.9351±0.0025& \textbf{0.7134±0.0030}
		\\
		\bottomrule
	\end{tabular}
    }
\end{table*}

\paragraph{Quantitative Results.}
The detailed results of the comparative evaluation are summarized in Table~\ref{tab:comparison_results}. As shown in Table~\ref{tab:comparison_results}, our model achieves the highest F1-score (0.8185), MCC (0.7134), precision (0.8227), and recall (0.8250), indicating its superior ability to balance predictive confidence and sensitivity. While ME-ACP slightly outperforms our model in AUC-ROC, it underperforms on F1 and MCC, suggesting potential overfitting or miscalibration.

\paragraph{Qualitative Insights.}
Figure~\ref{fig:tsne} presents a 2D t-SNE visualization of learned embeddings, showing clearer class separation in our model. Figure~\ref{fig:pr_curve} displays the micro-averaged precision-recall curve, where our method achieves the most stable and elevated profile across all recall levels.

\section{Conclusion}

We presented a novel adaptive multimodal transformer framework for antibody binding site prediction, systematically integrating sequence-based encodings, structural embeddings, and graph-derived biochemical features. By leveraging five biologically grounded modalities, our model employs an adaptive modality fusion mechanism that dynamically balances global informativeness, sample-specific variation, and class-aware semantics.
To further enhance representation capacity and robustness, we incorporate three key architectural advances: (i) a mixture-of-experts module for dynamic specialization; (ii) supervised contrastive learning to enforce class-level separation in the latent space; and (iii) stochastic weight averaging to stabilize training and improve generalization.
Extensive experiments on benchmark antibody-antigen datasets show that our method consistently outperforms competitive baselines across multiple evaluation metrics, including Precision, Recall, F1-score, AUC-ROC, and MCC. Detailed ablation studies confirm the complementary contributions of each modality and architectural component, validating the effectiveness of our multimodal design. This work highlights the potential of multimodal integration in antibody modeling and paves the way for future applications in immunoinformatics and therapeutic antibody discovery.

\bibliographystyle{unsrt}  
\bibliography{sample-base}  

\clearpage

\appendix

\section{Feature Representation in Bioinformatics}
\label{fea_rep}
\subsection{ESMC: Evolutionary Substitution Matrix Coding}
ESMC encodes protein sequences by leveraging evolutionary information derived from substitution matrices, capturing residue conservation and substitution patterns. This representation is particularly effective for tasks such as functional annotation and binding site prediction, where evolutionary conservation plays a critical role.

\subsection{One-hot Encoding}
One-hot encoding is a simple yet widely used method for representing amino acid sequences. Each residue is encoded as a binary vector of length 21, where a single bit is set to 1, corresponding to the amino acid type. Although straightforward, its lack of contextual and relational information limits its effectiveness for complex tasks.

\subsection{BLOSUM: Block Substitution Matrix}
The BLOSUM family of matrices, such as BLOSUM62~\cite{mount2008using}, provides a scoring system based on observed substitutions in conserved regions of proteins. These matrices incorporate evolutionary information and are often used in sequence alignment and similarity-based feature extraction, offering insights into residue-level functional importance.

\subsection{PyBioMed: Physicochemical Properties}
PyBioMed is a Python-based toolkit that extracts a wide range of physicochemical and structural features from protein sequences and structures, including hydrophobicity, charge, and secondary structure propensity~\cite{dong2018pybiomed}. These descriptors provide a rich feature set for downstream tasks, complementing sequence and evolutionary representations.

\subsection{Multi-view Learning in Bioinformatics}

Multi-view learning integrates complementary information from diverse feature representations, enabling more robust and accurate predictions in bioinformatics. For example, combining amino acid sequence features~\cite{clifford2022bepipred}, evolutionary profiles (e.g., BLOSUM and ESMC), and physicochemical properties~\cite{wang2023prediction} has consistently demonstrated superior performance in tasks such as protein function prediction and binding site identification. Recent advances in machine learning models, including Graph Neural Networks (GNNs) and Transformer-based architectures, have further enhanced multi-view learning by effectively capturing the relationships between different feature views.

\section{Evaluation Metrics}

We adopt five standard classification metrics to comprehensively assess model performance:

\begin{itemize}
	\item \textbf{Precision}: the proportion of true positives among predicted positives, defined as:
	\begin{equation}
		\text{Precision} = \frac{TP}{TP + FP},
	\end{equation}
	where $TP$ and $FP$ denote the number of true positive and false positive predictions, respectively.
	
	\item \textbf{Recall}: the proportion of true positives among actual positives, given by:
	\begin{equation}
		\text{Recall} = \frac{TP}{TP + FN},
	\end{equation}
	where $FN$ is the number of false negatives.
	
	\item \textbf{F1-score}: the harmonic mean of Precision and Recall, computed as:
	\begin{equation}
		\text{F1-score} = \frac{2 \cdot \text{Precision} \cdot \text{Recall}}{\text{Precision} + \text{Recall}}.
	\end{equation}
	
	\item \textbf{AUC-ROC}: the Area Under the Receiver Operating Characteristic Curve, which plots the true positive rate (TPR) against the false positive rate (FPR) at various thresholds. A higher AUC indicates better discrimination capability.
	
	\item \textbf{Matthews Correlation Coefficient (MCC)}: a balanced measure that considers true and false positives and negatives, particularly useful under class imbalance:
	\begin{equation}
		\text{MCC} = \frac{TP \cdot TN - FP \cdot FN}{\sqrt{(TP + FP)(TP + FN)(TN + FP)(TN + FN)}},
	\end{equation}
	where $TN$ denotes true negatives.
\end{itemize}

\end{document}